\documentclass[preprint,12pt]{elsarticle}

\usepackage{amsmath,amssymb,amsfonts}
\usepackage{algorithmic}
\usepackage{graphicx}
\usepackage{textcomp}
\usepackage{booktabs}
\usepackage{array}
\usepackage{tabularx}
\usepackage[T1]{fontenc}

\journal{}

\begin{document}

\begin{frontmatter}



\title{SPRMamba: Surgical Phase Recognition for Endoscopic Submucosal Dissection with Mamba}

\author[add1]{Xiangning Zhang\fnref{equal}}
\ead{zxnyyyyy@sjtu.edu.cn}
\author[add2]{Qingwei Zhang\fnref{equal}}
\ead{zhangqingwei@renji.com}
\author[add2]{Jinnan Chen\fnref{equal}}
\ead{23332@renji.com}
\author[add3]{Chengfeng Zhou}
\ead{joe_chief@hnu.edu.cn}
\author[add4]{Yaqi Wang}
\ead{wangyaqi@cuz.edu.cn}
\author[add1]{Zhengjie Zhang}
\ead{z876252209@sjtu.edu.cn}
\author[add2]{Xiaobo Li\corref{corresponding}}
\ead{lxb1969@sjtu.edu.cn}
\author[add1]{Dahong Qian\corref{corresponding}}
\ead{dahong.qian@sjtu.edu.cn}

\cortext[corresponding]{Corresponding author.}
\fntext[equal]{The three authors contribute equally to this work.}

\affiliation[add1]{organization={School of Biomedical Engineering},
            addressline={Shanghai Jiao Tong University}, 
            city={ShangHai},
            postcode={200000},
            country={China}}
\affiliation[add2]{organization={Division of Gastroenterology and Hepatology, Shanghai Institute of Digestive Disease, NHC Key Laboratory of Digestive Diseases, Renji Hospital},
	addressline={Shanghai Jiao tong University School of Medicine}, 
	city={ShangHai},
	postcode={200000}, 
	country={China}}
\affiliation[add3]{organization={Aier Institute of Digital Ophthalmology and Visual Science},
            addressline={Changsha Aier Eye Hospital}, 
            city={Changsha},
            postcode={410000},
            country={China}}
\affiliation[add4]{organization={College of Media Engineering},
            addressline={Communication University of Zhejiang}, 
            city={Hangzhou},
            postcode={310018},
            country={China}}
 
\begin{abstract}
Endoscopic Submucosal Dissection (ESD) is a minimally invasive procedure initially developed for early gastric cancer treatment and has expanded to address diverse gastrointestinal lesions. While computer-assisted surgery (CAS) systems enhance ESD precision and safety, their efficacy hinges on accurate real-time surgical phase recognition, a task complicated by ESD's inherent complexity, including heterogeneous lesion characteristics and dynamic tissue interactions. Existing video-based phase recognition algorithms, constrained by inefficient temporal context modeling, exhibit limited performance in capturing fine-grained phase transitions and long-range dependencies. To overcome these limitations, we propose SPRMamba, a novel framework integrating a Mamba-based architecture with a Scaled Residual TranMamba (SRTM) block to synergize long-term temporal modeling and localized detail extraction. SPRMamba further introduces the Hierarchical Sampling Strategy to optimize computational efficiency, enabling real-time processing critical for clinical deployment. Evaluated on the ESD385 dataset and the cholecystectomy benchmark Cholec80, SPRMamba achieves state-of-the-art performance (87.64\% accuracy on ESD385, +1.0\% over prior methods), demonstrating robust generalizability across surgical workflows. This advancement bridges the gap between computational efficiency and temporal sensitivity, offering a transformative tool for intraoperative guidance and skill assessment in ESD surgery. The code is accessible at https://github.com/Zxnyyyyy/SPRMamba.
\end{abstract}



\end{frontmatter}



\section{Introduction}
\label{sec:introduction}
Endoscopic Submucosal Dissection (ESD) is a minimally invasive procedure for treating early gastrointestinal cancers, offering advantages such as reduced trauma and faster recovery \cite{mccarty2020endoscopic}. However, the technical complexity of ESD, due to intricate lesion characteristics and the delicate gastrointestinal tract, limits its widespread adoption despite increasing demand for skilled endoscopists amid rising early cancer detection. Computer-assisted surgery (CAS) systems are an advanced medical technology that can significantly enhance ESD efficiency and reduce the risk of complications \cite{maier2017surgical}. A critical component of CAS is surgical phase recognition (SPR), which enables real-time monitoring, procedural optimization, and surgical education by analyzing video data \cite{twinanda2016endonet, vercauteren2019cai4cai}. Therefore, developing efficient and accurate video-based SPR algorithms is essential to meet the demands of modern surgical practice and education.

\begin{figure*}[!t]
	\centerline{\includegraphics[width=\textwidth]{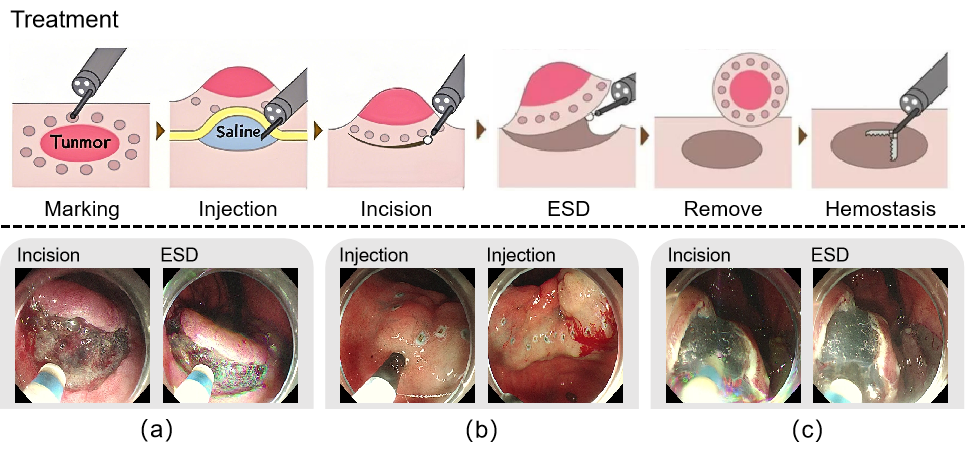}}
	\caption{The first part shows the conventional steps of ESD. The second part gives examples of various challenges for automated phase recognition from ESD videos. The text in the top-left corner of each image indicates which phase it belongs to. (a) limited inter-phase variance, (b) significant intra-phase variance, and (c) transition phase similarity.}
	\label{background}
\end{figure*}

Limited inter-phase variance (subtle visual distinctions between phases see Fig.\ref{background}(a)) and high intra-phase variance (diverse tool motions within a single phase see Fig.\ref{background}(b)) are the most critical challenges for automatically recognizing surgical phases from video. Recent studies (e.g., TMRNet \cite{jin2021temporal}, TeCNO \cite{czempiel2020tecno}, Trans-SVNet \cite{gao2021trans}) solved this issue by capturing the long-term frame-wise relation between the current frame and previous frames. However, these methods falter in ESD's unique environment. Because of TCNs \cite{farha2019ms} lose fine-grained temporal details due to fixed convolutional kernels, while Transformers incur prohibitive computational costs for long, uncut ESD videos. Moreover, existing frameworks prioritize short-term dependencies, neglecting ESD's heterogeneous phase durations. These limitations exacerbate misclassifications, particularly during transitions between visually similar phases (e.g., incision vs. ESD, see Fig.\ref{background}(c)) \cite{huang2023experimental, cao2023intelligent, furube2024automated}.

Recently, a state-space model called Mamba has demonstrated substantial potential in modeling temporal contexts. Unlike traditional temporal models, the Mamba model shows higher robustness and accuracy in dealing with complex surgical phase transitions while avoiding the problem of the Transformer's high computational complexity. Making Mamba well suited for modeling the temporal context in ESD videos. To fully utilize Mamba's temporal modeling ability, we propose SPRMamba, a novel framework that integrates Mamba’s linear-complexity state-space modeling with a lightweight Transformer to capture both long-range dependencies and fine-grained temporal features. Furthermore, according to \cite{bahrami2023much}, allowing the model to operate on the complete input sequence is more beneficial compared to only accessing a subset of the input. Therefore, considering the quadratic complexity of the transformer and the high memory usage of $\mathcal O(L^{2})$ (where L represents the input sequence length), it is inappropriate to directly apply the transformer to uncut surgical videos. To address this issue, SPRMamba employs a hierarchical sampling Strategy (Long-range and Window Sampling) to reduce computational overhead, support the online phase, and the LSTContext modules to aggregate multi-scale temporal contexts. Our main contributions can be summarized as follows:

\begin{itemize}
  \item We propose a novel surgical phase recognition framework (SPRMamba), leveraging the Scaled Residual TranMamba (SRTM) module to efficiently model short-term and long-term temporal contexts. To the best of my knowledge, this is the first research to develop a Mamba-based surgical phase recognition model.
  \item We introduce Hierarchical Sampling Strategy with Window and Long-range sampling to reduce computational burden and support online phase recognition. 
  \item Conducting qualitative and quantitative experiments on ESD385 and Cholec80 datasets, demonstrating significant improvements over existing techniques.
\end{itemize}

\section{Related Work}
\label{sec:related work}
\subsection{Video understanding}
Effective spatiotemporal modeling is foundational for video understanding, yet existing approaches face critical trade-offs in medical applications like ESD. Early 3D-CNNs \cite{ji20123d} jointly modeled spatial and temporal features but incurred prohibitive computational costs (\(O(T \cdot H \cdot W)\)) for hour-long ESD videos. To address computational limitations, Temporal Convolutional Networks (TCNs) \cite{bai2018empirical} emerged, which focus on capturing temporal dependencies through one-dimensional convolutions. While TCNs are more efficient and capable of handling longer video sequences, they struggle to model complex action variations and capture the fine-grained details needed for surgical phase recognition. Transformer-based models have gained attention in recent years for their ability to model long-range dependencies using self-attention mechanisms \cite{girdhar2019video}. While Transformers achieve state-of-the-art performance in video understanding tasks by capturing spatiotemporal relationships across frames, but introduce quadratic complexity (\(O(L^2)\)), limiting real-time deployment. Despite these advancements, the complexity of ESD surgery poses unique challenges for current methods in video understanding. Surgical phases often involve intricate and fine-grained transitions that are difficult to capture with traditional models. In addition, effectively modeling short-term and long-term spatiotemporal dependencies in long videos while maintaining controllable computational costs remains a major challenge.

\subsection{Surgical Phase Recognition}
Research in the field of Surgical Phase Recognition mainly focuses on automatically detecting and classifying different phases of surgery by analyzing surgical videos. Early surgical phase recognition methods relied heavily on hand-designed features and statistical learning methods \cite{zappella2013surgical, blum2010modeling, lalys2011framework, dergachyova2016automatic, twinanda2016endonet, tao2013surgical, quellec2014real, lea2015improved}. However, these methods are limited by empirical design features and usually rely on predefined dependencies, making it difficult to accurately capture subtle motion features with strongly nonlinear dynamics, and their performance suffers greatly. Twinanda et al. \cite{twinanda2016endonet} proposed the EndoNet model, a CNN-based approach that automatically extracts features and performs phase classification from laparoscopic surgical videos. This approach opened the way for SPR research using deep learning. Since then, researchers have sought to enhance SPR by incorporating more advanced temporal modeling techniques. For instance, long short-term memory (LSTM) networks have been widely adopted in models like PhaseNet \cite{twinanda2016single}, EndoLSTM \cite{twinanda2017vision}, and SV-RCNet \cite{jin2017sv}, enabling the capture of both spatial and temporal dependencies. These methods, by combining deep residual networks (ResNet) with LSTM modules, have shown promising improvements in recognizing surgical phases by effectively modeling temporal series data. However, these methods are computationally expensive when dealing with long time series and prone to misclassification when dealing with complex phase transitions. To address these limitations, Transformer-based architectures have been introduced into SPR due to their capability of capturing long-range temporal dependencies. For example, Czempiel et al. \cite{czempiel2021opera} proposed a Transformer-based surgical phase recognition method, which significantly improves the model's performance in complex surgical scenarios through the self-attention mechanism. More recently, Graph Neural Networks (GNNs) have been employed in SPR, with studies such as Padoy et al. \cite{padoy2019machine} demonstrating their effectiveness in modeling complex spatiotemporal relationships between surgical tools and anatomical structures but added architectural complexity, hindering real-time use. These limitations underscore three unresolved challenges: (1) efficient long-sequence modeling, (2) fine-grained temporal resolution, and (3) real-time inference—critical for ESD's clinical demands. Therefore, this paper aims to construct a novel ESD surgical phase recognition model that is capable of maintaining high accuracy while reducing computational complexity.

\subsection{State Space Models}
Recently, State-Space Models (SSMs) have been proven to have transformer-level performance in capturing long sequences in the field of natural language processing \cite{gu2021efficiently, gu2023Mamba}. \cite{gu2021efficiently} introduced a new model for Structured State-Space Sequences (S4), specifically designed to model long-range dependencies exhibiting the well-established property of scaling linearly with sequence length. Based on this, \cite{smith2022simplified} proposed an advanced layer called S5, which integrates MIMO SSM and efficient parallel scanning into the S4 architecture. This development aims to overcome the limitations of SSMs and improve their efficiency. In addition, \cite{fu2022hungry} contributed a novel SSM layer H3, significantly narrowing the performance gap between SSM and transformer-based attention in language modeling. \cite{mehta2022long} Expand the S4 model by introducing additional gating units in the gated state space layer to enhance its expressiveness. More recently, \cite{gu2023Mamba} developed a universal language model called Mamba by introducing a data-dependent SSM layer and a selection mechanism using parallel scanning. Compared to transformers based on quadratic complexity attention, Mamba excels at handling long sequences with linear complexity. 
 
In the field of vision, \cite{zhu2024vision} proposed Visual Mamba (Vim), which combines position encoding and bi-directional scanning to efficiently capture the global context of an image. Pioneered the application of Mamba in vision tasks. Li et al. \cite{li2024videoMamba} constructed a generic framework called Video Mamba Suite to develop, validate, and analyze Mamba's performance in video understanding. Besides, the great potential of Mamba has inspired a series of works \cite{liu2024vMambavisualstatespace, yang2024plainMamba, pei2024efficientvMamba, huang2024localMamba, chen2024mim}, which demonstrated that Mamba has better performance and higher GPU efficiency than Transformer on visual downstream tasks such as semantic segmentation and video understanding. However, no prior work applies SSMs to surgical phase recognition, particularly in ESD's challenging environment of low inter-phase variance and extreme temporal imbalance. In our work, we integrate Mamba into the surgical phase recognition model to efficiently capture the temporal context in ESD video.

\section{Methodology}
\label{sec:methodology}
\subsection{Preliminaries}
State Space Models (SSMs) are typically considered linear time-invariant systems that map a 1-D function or sequence \( x(t) \in \mathbb{R} \) \(\to\) \( y(t) \in \mathbb{R} \) through a hidden state \( h(t) \in \mathbb{R}^N \). These systems are commonly represented by linear ordinary differential equations (ODEs) with \( A \in \mathbb{R}^{N \times N} \) as the evolution parameter, and \( B \in \mathbb{R}^{N \times 1} \), \( C \in \mathbb{R}^{1 \times N} \) as projection parameters.
\begin{equation}h'(t)=Ah(t)+Bx(t), y(t)=Ch(t).\label{eq1}\end{equation}

Structured State Space Sequence Models (S4) and Mamba are discrete versions of continuous systems, which include a time scale parameter \( \Delta \) that converts continuous parameters \( A \) and \( B \) into discrete parameters $\bar{A}$, $\bar{B}$. \(A\) common conversion method is Zero-Order Hold (ZOH), defined as follows:
\begin{equation}\bar{A}=\exp(\Delta A), \bar{B}=(\Delta A)^{-1}(\exp(\Delta A)-I)\cdot\Delta{B}.\label{eq2}\end{equation}

After the discretization of $\bar{A}$, $\bar{B}$, Eq.(\ref{eq1}) can be expressed as discrete parameters:
\begin{equation}h_{t}=\bar{A}h_{t-1}+\bar{B}x_{t}, y_{t}=Ch_{t}.\label{eq3}\end{equation}

Finally, for an input sequence of size $T$, the output $y$ is calculated using a global convolution operation with a convolution kernel $\bar{K}$
\begin{equation}\bar{K}=({C\bar{B}}, {C\bar{AB}} ,..., {C\bar{A}^{M-1}\bar{B}}), {y}={x*\bar{K}},\label{eq4}\end{equation}
where M is the length of the input sequence x, and \({\bar{K}} \in \mathbb{R}^M \) is a structured convolutional kernel.

\begin{figure*}[!t]
	\centerline{\includegraphics[width=\textwidth]{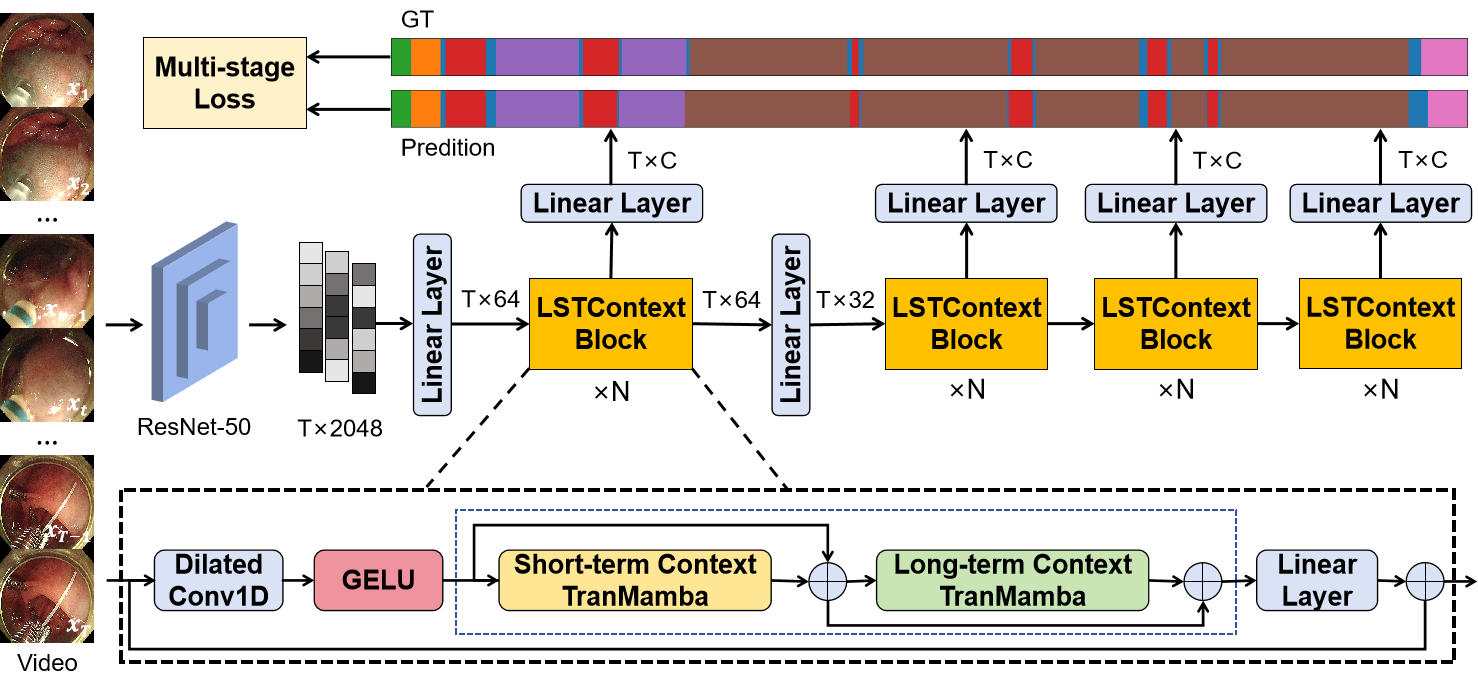}}
	\caption{A schematic overview of the proposed SPRMamba architecture, which consists of a ResNet-50, and four LSTContext blocks (top).}
	\label{framework}
\end{figure*}

\subsection{Overview}
The architecture of the proposed SPRMamba is shown in Fig.\ref{framework}. For a video \( V \in \mathbb{R}^{L \times C \times H \times W} \) of length \({L}\), we first extract the spatial frame-level feature sequence \( {F} \in \mathbb{R}^{L \times D} \) from a fixed ResNet-50 \cite{he2016deep}, where \(D=2048\) is the spatial dimension, and then SPHMamba uses a linear layer to reduce the feature dimension to 64. As in previous works, we repeated each LSTContext block \(N\) times, where the dilation factor of the dilation convolution was increased in each layer. In practice, we use \(N = 10\) and we evaluate the impact of \(N\) in Section \ref{sec:experiments}. After the first \(N\) layers of the LTContext block, we use an additional linear layer to further reduce the dimensionality \(D\) to 32. The dimensionality reduction method reduces the number of parameters from 2.49 million to 1.23 million without degrading the accuracy. We also use an additional Conv1D followed by a softmax layer to generate the frame-level class probabilities \( {P} \in \mathbb{R}^{L \times C} \). 

We proceed with three additional stages, each consisting of \(N\) layers of LSTContext blocks. At the beginning of each stage, we reset the dilation factor of the temporal convolution to 1 and compute the frame-level class probabilities \(P \in \mathbb{R}^{T \times C}\) after each stage, resulting in a multi-stage loss. We use cross-entropy loss and the mean squared error smoothing loss introduced by \cite{farha2019ms} for supervised training. Inspired by \cite{yi2021asformer}, we apply the cross-attention to the LSTContext blocks in stages 2 to 4, instead of using feature \({F}\) for queries and keys for window and long-term context samples, but instead using predictive \({P}\). We evaluate the impact of the number of stages in Section \ref{sec:experiments}.

\subsection{Scale Residual TranMamba}
\begin{figure*}[!t]
	\centerline{\includegraphics[width=0.8\textwidth]{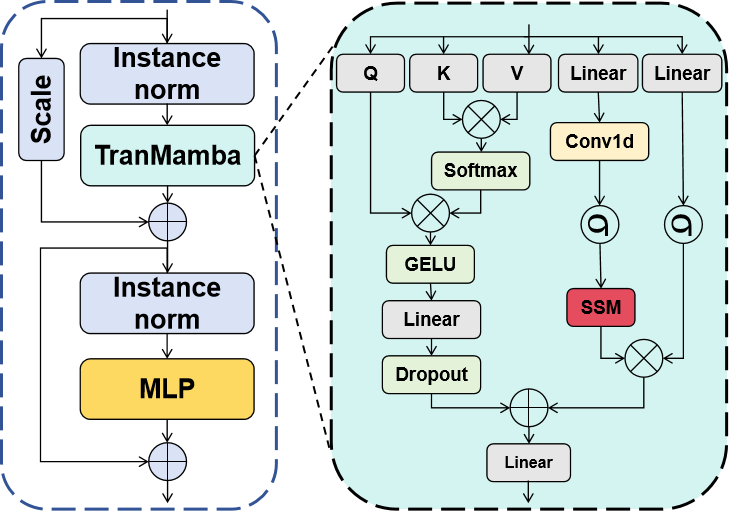}}
	\caption{A schematic overview of the SRTM module, which is composed of the TranMamba and its main components.}
	\label{SRTM}
\end{figure*}
For the surgical phase recognition task, temporal information plays a crucial role in accurate recognition. However, due to the complexity of ESD surgeries, existing surgical phase recognition algorithms based on traditional temporal models are difficult to apply directly to ESD data. We aim to explore a simple and effective structure combining Mamba, which simultaneously models both long-term and short-term temporal contexts. A straightforward approach is to combine the short-term temporal context modeled by the transformer with the long-term temporal context modeled by Mamba. Therefore, we propose the Scaled Residual TranMamba module, as shown in Fig.\ref{SRTM}.

Given the input feature ${F_{in}} \in \mathbb{R}^{L \times C}$, the SRTM module first applies Instance norm \cite{ulyanov2016instance}. Then, it uses the TranMamba module to capture long-term and short-term temporal context, thereby generating ${F_{ls}} \in \mathbb{R}^{L \times C}$. In addition, to obtain more comprehensive context information, the fusion of ${F_{in}}$ and ${F_{ls}}$ was achieved through scale residual connections. The fused feature is followed by normalization utilizing Instance Norm, and then an MLP to learn deeper features. The entire process can be described as follows:
\begin{equation}{F}=\beta{F_{in}}+TranMamba(IN({F_{in}}))\label{eq5}\end{equation}
\begin{equation}{F_{out}}=MLP(IN({F}))\label{eq6}\end{equation}

Specifically, there are three branches in the TranMamba module. The first branch takes the first quarter portion of the input feature ${F_{in}}$ along the channel dimension as input ${F_{b1}} \in \mathbb{R}^{L \times \frac{C}{4}}$, then expands the dimension to $\lambda$${C}$ by a linear transformation, and then activates it using the SiLU \cite{ramachandran2017searching} function. The second branch in the TranMamba module has inputs similar to the first branch. It takes the second quarter of the input features ${F_{in}}$ along the channel dimension as input ${F_{b2}} \in \mathbb{R}^{L \times \frac{C}{4}}$. Subsequently, these features are sequentially expanded through the dimensional expansion of the linear layer, Conv1d layer, SSM, and LayerNorm. Afterward, the features extracted from both branches are fused using the Hadamard product, which is designed to capture the long-term temporal context information in this way \cite {gu2023Mamba}. The third branch takes the last half of the input features ${F_{in}}$ along the channel dimension as input ${F_{b3}} \in \mathbb{R}^{L \times \frac{C}{2}}$. Subsequently, ${F_{b3}}$ is input into the Attention layer and activated using the GELU function \cite{hendrycks2016gaussian}, aiming to capture short-term temporal context features. Finally, the short-term temporal features achieved from the third branch will be fused with the previously captured long-term temporal features via a linear mapping to obtain feature ${F_{ls}}$. The entire process takes the following form:

\begin{equation}{F_{1}}=SiLU(Linear({F_{b1}}))\label{eq7}\end{equation}
\begin{equation}{F_{2}}=LN(SSM(Conv1d(Linear({F_{b2}}))))\label{eq8}\end{equation}
\begin{equation}Attention(Q,K,V)=softmax(\frac{QK^{T}}{\sqrt{K}}V)\label{eq9}\end{equation}
\begin{equation}{F_{3}}=Dropout(Linear(GELU(Attention({F_{b3}}))))\label{eq10}\end{equation}
\begin{equation}{F_{ls}}=Linear({F_{1}}\odot{F_{2}}+{F_{3}})\label{eq11}\end{equation}

where $\odot$ represents Hadamard product, ${Q,K,V \in \mathbb{R}^{L \times D}}$ are linearly transformed form ${F_{b3}}$ 

\subsection{Hierarchical Sampling Strateg}
\begin{figure*}[!t]
	\centerline{\includegraphics[width=\textwidth]{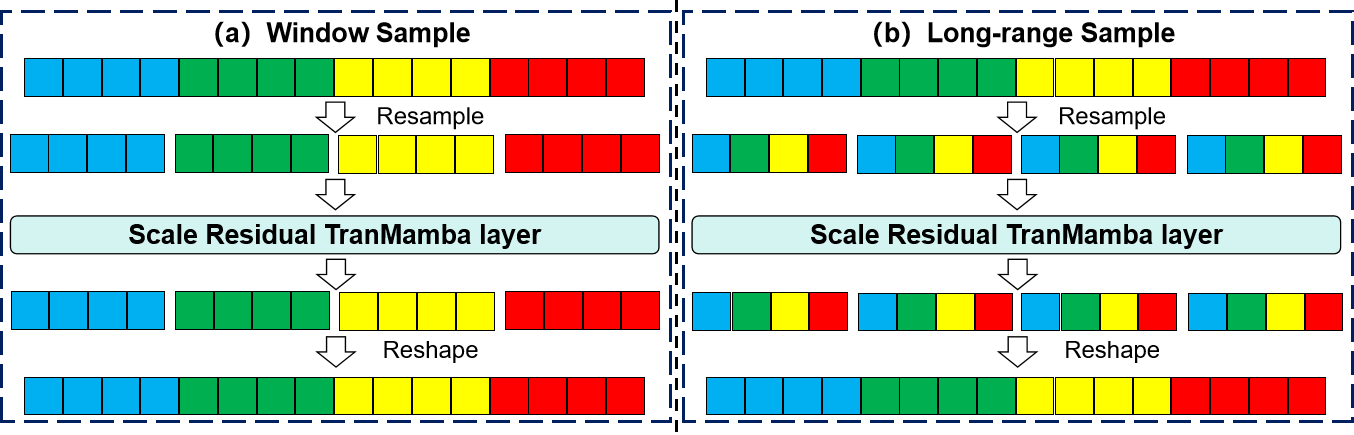}}
	\caption{The Hierarchical Sampling Strateg is illustrated with a window of size 4. (a) For window sampling, the sequence is partitioned into small windows, and the SRTM is computed for each window. (b) For long-term sampling, the sequence is reordered such that the SRTM is computed over the entire, but sparsely sampled sequence. After the SRTM computation, the output is reordered to preserve the original sequence order.}
	\label{HSS}
\end{figure*}

Due to the quadratic complexity of self-attention blocks, it is impractical to apply attention to long sequences of untrimmed videos. This is because the L of the video sequence is very large, so we need to resample it to achieve modeling of long-term and local temporal contexts, as shown in Fig.\ref{HSS}.

\textbf{Temporal Window Sample.} For temporal window sampling, we divide the sequence into non-overlapping windows of size \(W\). Fig.\ref{HSS} illustrates the case of \(W=4\). Given that different tasks may have different time-dependent ranges, W is task-specific. We used \(W=64\) in practice. The impact of W was evaluated in Section \ref{sec:experiments}. Instead of modeling the temporal context over the entire sequence of length \(T\), we model the temporal context ${\frac{T}{W}}$ times, where each ${F_{b1}} \in \mathbb{R}^{W \times \frac{C}{4}}$, ${F_{b2}} \in \mathbb{R}^{W \times \frac{C}{4}}$, ${F_{b3}} \in \mathbb{R}^{W \times \frac{C}{2}}$ corresponds to each window. 

\textbf{Temporal Long-range Sample.} For the temporal long-range sample, we sample the input every \(G\) to divide \(G\) non-overlapping sequences and model the temporal context information of each sequence. In ${G=4}$ shown in Fig.\ref{HSS} b, we model the temporal context information of each of the four downsampled sequences. In general, we model the temporal contexts for \(G\) ${F_{b3}} \in \mathbb{R}^{W \times \frac{C}{2}}$ where the ${F_{b1}}$ and ${F_{b2}}$ are the same, $\textit{i.e.}$, ${F_{b1},F_{b2}} \in \mathbb{R}^{W \times \frac{C}{4}}$. The parameter \(G\) provides the flexibility to adjust the sparsity based on the available memory budget, e.g., \(G=1\) corresponds to the case where the attention is applied to the entire sequence. In practice, we use \(G=64\) and evaluate the impact of \(G\) in Section \ref{sec:experiments}.

\textbf{LSTContext Block.} The top of Fig.\ref{framework} illustrate the entire LSTContext block. As in previous work \cite{czempiel2020tecno}, we use a 1D dilated convolution with kernel size 3. This is because the dilated convolution increases the receptive field without the need to increase the parameter number by increasing the kernel size. Where the dilation factor for each layer increases by a factor of 2, and the receptive field exponentially expands as the number of layers increases. Therefore, with a few parameters, we achieved a significantly larger receptive field in the temporal sequence, which mitigated model overfitting and effectively promoted the accuracy of surgical phase recognition. The dilated convolution is followed by a Gaussian Error Linear Unit (GELU). In the LSTContext block, we first use the window sample and then the long-range sample, as shown in Fig.\ref{HSS}. Finally, we use a linear layer with residual connectivity to output the features for each frame, ${F} \in \mathbb{R}^{L \times D}$

\section{Experiments and Results Analysis}
\label{sec:experiments}
In this section, we first describe the dataset used in our study, as well as the detailed experimental setup and evaluation metrics employed. Subsequently, we validated the effectiveness of our proposed method in ESD and cholecystectomy by comparing it with the SOTA method and conducting ablation experiments.
\subsection{ESD385 Dataset}
\begin{figure*}[!t]
    \centerline{\includegraphics[width=\textwidth]{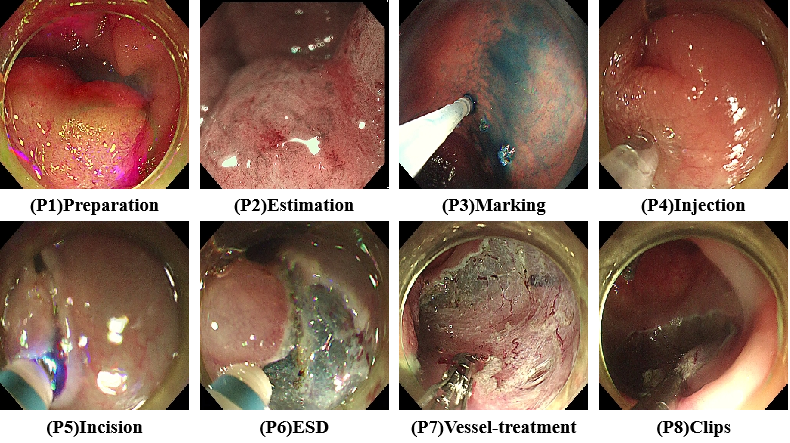}}
	\caption{Illustration of 8 surgical phases (P1-P8) annotated in the ESD385 dataset.}
	\label{demo}
\end{figure*}
\textbf{ESD385} 
retrospectively selected patients who underwent ESD from August 16, 2023, to January 8, 2024, in the endoscopy unit of the Department of Gastroenterology, Renji Hospital, Shanghai Jiao Tong University. A total of 385 videos of ESD procedures were collected. All procedures were performed by experienced endoscopists. The instruments used for ESD procedures included: gastroscope GIF-Q260, GIF-H260, GIF-HQ290, mucosal incision knife KD-612L/U, KD-655L/U, KD-620UR, endoscopic water pump OFP-2, mucosal injection needle NM-400L-0423, NM-400U-0423, thermal biopsy forceps HDBF-2.4-230-S, hemostatic forceps FD-410LR, FD-412LR, soft tissue clips ROCC-D-26-195-C, ROCC-D-26-235-C, hemostatic clips M00521242, methylene blue injection, indigo rouge solution. All endoscopic surgical videos were recorded using Image Management Hub, IMH-200, and Olympus and stored in MP4 format. The video resolution was 1920 × 1080 with a frame rate of 50 fps. After collection, all videos were anonymized to remove identifiable patient information, procedure timestamps, and other identifying details. This work was approved by the Renji Hospital affiliated with the Shanghai Jiao Tong University School of Medicine ethics committee with number RA-2024-457(2024.3.20).
\begin{table}[ht]
\caption{Distribution of annotations in the ESD385 dataset for phase recognition.}
\label{tab3}
\centering
\setlength{\tabcolsep}{12pt}
\begin{tabular}{c|c|c|c}
\toprule
 & Train & Val & Test \\
\hline
Preparation & 91739 & 14193 & 14323 \\
\hline
Estimation & 43198 & 5822 & 16860 \\ 
\hline
Marking & 15585 & 2820 & 6583 \\
\hline
Injection & 44464 & 9886 & 15831 \\
\hline
Incision & 59695 & 11917 & 22181 \\
\hline
ESD & 174201 & 41807 & 75968 \\
\hline
Vessel-treatment & 44317 & 4865 & 14707 \\
\hline
Clips & 10817 & 3985 & 4748 \\
\hline
Total & 484016 & 95295 & 171201 \\
\bottomrule
\end{tabular}
\end{table}

The ESD video is divided into eight surgical phases, including (1) Preparation; (2) Estimation; (3) Marking; (4) Injection; (5) Incision; (6) ESD; (7) Vessel-treatment; and (8) Clips(portion). Samples are shown in Fig.\ref{demo}. Phase (1) Preparation included elements unrelated to the ESD procedure, such as gastric insufflation and device replacement. Four endoscopists independently annotated the video, labeling each frame in the video. After the initial annotation, quality control was performed by two additional experienced endoscopists. Any uncertainties that arose during the quality control process were resolved through collaborative discussions among the six experts. The number of annotated frames in the dataset varied at each phase, with the ESD phase occupying most of the procedure time, which is the most important and skill-demanding phase of ESD. Detailed statistics for each phase are shown in Table \ref{tab4}. Overall, a total of 484016 frames, 95295, and 171201 frames were annotated for training, validation, and verification, respectively.

\textbf{Cholec80} 
is a large-scale surgical benchmark dataset containing 80 cholecystectomy videos performed by 13 surgeons. These videos are recorded at 25 frames per second, with each frame having a resolution of either 1920×1080 or 854×480 pixels. Each frame in the videos is annotated with one of seven surgical phases: Preparation (P1), Calot triangle dissection (P2), Clipping and cutting (P3), Gallbladder dissection (P4), Gallbladder packaging (P5), Cleaning and coagulation (P6), and Gallbladder retraction (P7). Additionally, each frame includes annotations for seven types of surgical tools, including Grasper, Bipolar, Hook, Clipper, Scissors, Irrigator, and Specimen bag. We strictly followed the experimental protocol used in previous studies \cite{twinanda2016endonet,jin2017sv,jin2021temporal}, dividing the dataset into two equal subsets: the first 40 videos for training and the remaining 40 videos for testing, with 8 videos from the training set selected for validation.

\subsection{Experiments Design}
\textbf{Evalutation Metrics.} For surgical phase recognition, we employed four commonly used evaluation metrics to quantitatively assess model performance. These metrics include Precision, Recall, Jaccard Index, and Accuracy. Accuracy is defined as the percentage of frames across the entire video correctly predicted to be in their ground truth phase. Given the imbalanced phase presented in the video, Precision, Recall, and Jaccard Index refer to phase-level evaluations, calculated within each phase and then averaged across all phases. For the Cholec80 dataset, our evaluation protocol adheres to the benchmark standards established (Relaxed metric) in prior work \cite{twinanda2016single} and aligns with recent approaches \cite{twinanda2016endonet,jin2017sv,jin2020multi,jin2021temporal}, considering a result accurate if the temporal discrepancy from the ground truth remains within a 10-second threshold. Furthermore, we consistently utilized the standard metric in other experiments.

\textbf{Implementation details.} Our approach is implemented in Python using the PyTorch framework, and training is conducted on a workstation equipped with an NVIDIA RTX 4090. In the first stage, we use a ResNet-50 model pre-trained on ImageNet-22K \cite{he2016deep}. We then fine-tune the model on our data. To ensure a fair comparison with SOTA methods, we downsample all videos to 1 fps, which is also the approach used in previous works \cite{jin2020multi,jin2021temporal}. This operation has additional benefits, including enriching temporal information and saving memory. During training, image frames are further downsampled from the original resolutions of 1920 $\times$ 1080 and 854 $\times$ 480 to a resolution of 250 $\times$ 250 pixels to further reduce memory usage. Data augmentation is performed through 224 $\times$ 224 cropping, flipping, and random mirroring to expand the training dataset. The ResNet-50 model is fine-tuned with a batch size of 160 images. In the second stage, LSTContext is trained end-to-end from scratch. The W and G for each task in the LSTContext module are initially chosen based on estimates of task average durations on the training dataset and are then fine-tuned through ablation studies. For both stages, we adopted the same experimental setup as the literature \cite{he2016deep}, using the AdamW optimizer \cite{loshchilov2017decoupled} with a weight decay of 1e-5 and approximately 200 iterations. A cosine learning rate scheduler \cite{loshchilov2016sgdr} is used, with 40 epochs of linear warm-up and an initial learning rate of 5e-4.

\subsection{Comparison with State-of-the-Art Methods}
\subsubsection{Result on the ESD385 Dataset}
On the ESD385 dataset, we compared our proposed method with the SOTA method. This includes the method proposed by Furube et al. \cite{furube2024automated}, which fine-tunes ResNet-50 as a feature extractor and then uses MS-TCN for hierarchical prediction refinement for surgical phase recognition. The method proposed by Cao et al. \cite{cao2023intelligent}. An Intelligent Surgical Workflow Recognition Suite for ESD is based on Trans-SVNet \cite{gao2021trans} implementation. In addition, we compared our method with SV-RCNet \cite{jin2017sv} and SAHC \cite{ding2022exploring}, two SOTA methods for cholecystectomy surgical phase recognition. 
\begin{table}[!h]
\caption{Comparison with the SOTA methods on the ESD385 dataset.}
\label{tab1}
\centering
\setlength{\tabcolsep}{3pt}
\resizebox{\textwidth}{!}{
\begin{tabular}{c|c|c|c|c|c|c}
\toprule
Method & Accuracy(\%) & Precision(\%) & Recall(\%) & Jaccard(\%) & FLOPs & Params \\
\hline
ResNet-50 \cite{he2016deep} & 72.31 & 78.75 & 66.63 & 55.39 & 4.1G & 24.56M \\
\hline
SV-RCNet \cite{jin2017sv} & 75.58$\pm$13.46 & 81.27$\pm$17.23 & 70.78$\pm$17.46 & 60.12$\pm$17.77 & 41.1G & 28.76M \\
\hline
SAHC \cite{ding2022exploring} & 86.64$\pm$10.63 & 86.35$\pm$18.07 & 83.75$\pm$17.55 & 75.14$\pm$18.20 & 9.5G & 26.27M \\
\hline
Furube et at. \cite{furube2024automated} & 84.79$\pm$11.58 & 84.85$\pm$18.75 & 82.11$\pm$17.52 & 72.60$\pm$18.25 & 4.5G & 24.69M \\
\hline
AI-Endo \cite{cao2023intelligent} & 83.38$\pm$12.09 & 84.74$\pm$17.35 & 82.17$\pm$17.23 & 72.09$\pm$17.23 & 5.7G & 24.72M\\
\hline
\textbf{SPRMamba(Ours)} & \textbf{87.70$\pm$11.03} & \textbf{86.84$\pm$16.07} & \textbf{86.89$\pm$15.64} & \textbf{77.74$\pm$17.60} & 7.5G & 25.42M \\
\bottomrule
\end{tabular}
}
\end{table}

The comparative results are presented in Table \ref{tab1}. We found that applying existing surgical phase recognition methods directly to ESD leads to performance degradation because ESD surgery has more complex workflows compared to traditional surgeries. In the case of image classification using ResNet-50 alone, the average accuracy is 72.31\%, the average precision is 78.75\%, the average recall is 66.63\%, and the average Jaccard is 55.39\%. After including the SRTM module and Hierarchical Sampling Strateg, the present method achieves an average accuracy of 87.64\%, an average precision of 86.72\%, an average recall of 86.76\%, and an average Jaccard of 77.51\%, increasing the average accuracy from 72.31\% to 87.64\%. These results demonstrate the effectiveness of the method for automatic phase recognition. Compared to the SOTA method, our method outperforms the second-best method in all four metrics by 1.00\%, 0.37\%, 3.01\%, and 2.37\%, respectively. Furthermore, to demonstrate the algorithm efficiency of the proposed method. We also compare the number of parameters and computational cost of our method with the SOTA method, which is reported in Table \ref{tab1}. "Params" denotes the number of parameters. Our proposed method outperforms SAHC in terms of fewer parameters. Compared to SAHC, our model will reduce 2.0G FLOPS.
\begin{figure*}[!t]
	\centerline{\includegraphics[width=\textwidth]{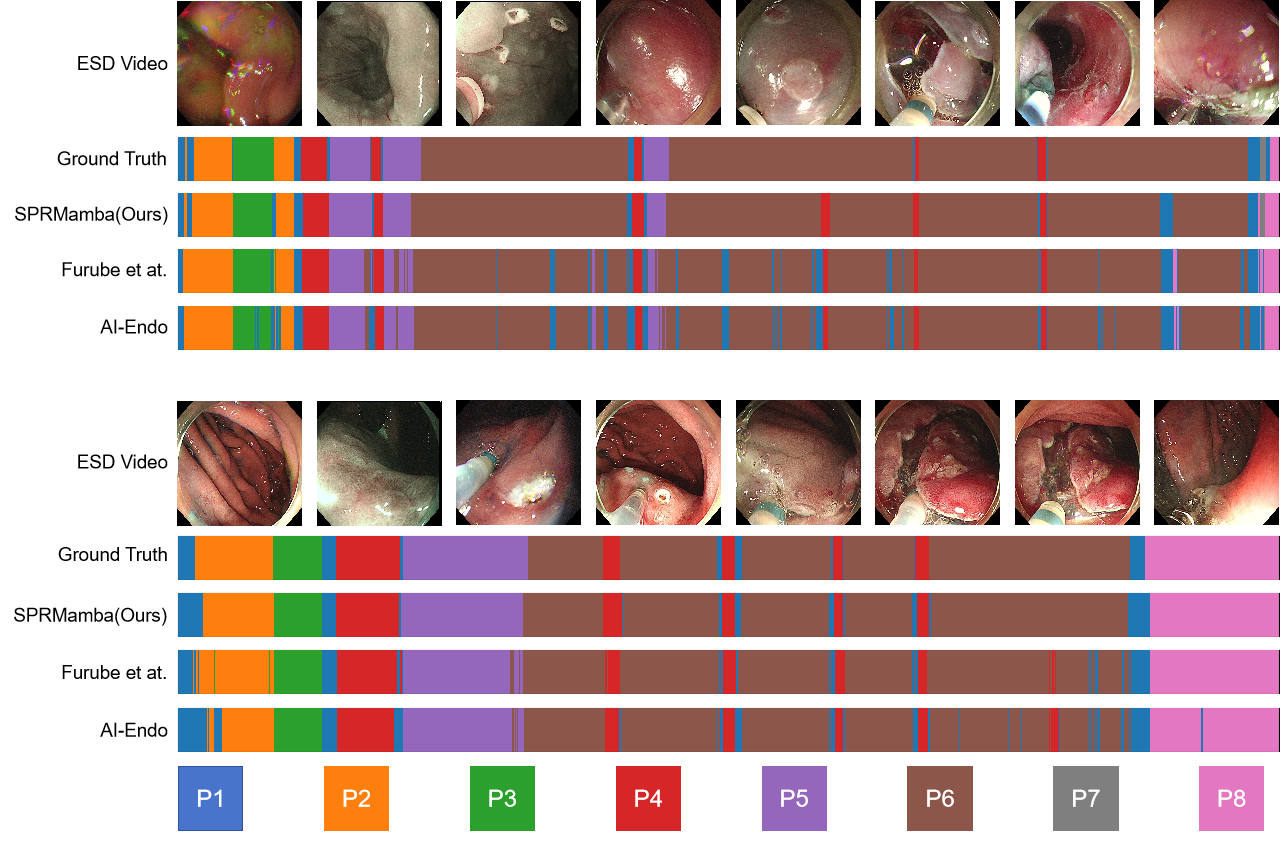}}
	\caption{Qualitative comparisons with some other methods on ESD385 dataset. The following four rows show the ground-truth labels, the predictions by the proposed SPRMamba, the predictions by Furube et at., and the predictions by AI-Endo. P1 to P8 indicate the phase label.}
	\label{fig4}
\end{figure*}

In Fig.\ref{fig4}, we visualized the predictions of two ESD videos to qualitatively show the improvement in surgical phase recognition. The results highlight that SPRMamba obtains consistent and smooth predictions not only within one phase but also for the often ambiguous phase transitions. Compared with AI-Endo and Furube et at., SPRMamba can perform accurate phase recognition even for phases with short durations, such as phase Preparation and Clips. Finally, SPRMamba shows robustness because case 2 lacks phase Clips. However, the performance of our model does not deteriorate.

\subsubsection{Result on the Cholec80 Dataset}
In the Cholec80 dataset, we conducted a performance comparison of our methods with state-of-the-art (SOTA) methods, including LAST \cite{tao2023last}, SV-RCNet \cite{jin2017sv}, TeCNO \cite{czempiel2020tecno}, TMRNet \cite{jin2021temporal}, Trans-SVNet \cite{gao2021trans}, SAHC \cite{ding2022exploring}, SKiT \cite{liu2023skit}, and LoViT \cite{liu2025lovit}. Please note that we re-implemented TeCNO and Trans-SVNet using the model weights provided in the original manuscript. The results of the other state-of-the-art methods were extracted verbatim from their respective published works. 
\begin{table}[ht]
\caption{Comparison with the SOTA methods on the Cholec80 dataset. Note that the ‘+’ denotes methods based on multi-task learning that require extra tool labels.}
\label{tab2}
\centering
\setlength{\tabcolsep}{3pt}
\resizebox{\textwidth}{!}{
\begin{tabular}{c|c|c|c|c|c}
\toprule
Method & Relaxed metric & Accuracy(\%) & Precision(\%) & Recall(\%) & Jaccard(\%) \\
\hline
LAST \cite{tao2023last}$^{+}$ & \checkmark & 93.12$\pm$4.71 & 89.25$\pm$5.49 & 90.10$\pm$5.45 & 81.11$\pm$7.62 \\
\hline
SV-RCNet \cite{jin2017sv} & \checkmark & 85.30$\pm$7.30 & 80.70$\pm$7.00 & 83.50$\pm$7.50 & - \\
\hline
TeCNO \cite{czempiel2020tecno} & \checkmark & 90.17$\pm$6.91 & 88.13$\pm$5.29 & 87.07$\pm$6.62 & 76.01$\pm$7.78 \\
\hline
TMRNet \cite{jin2021temporal} & \checkmark & 90.10$\pm$7.60 & 90.30$\pm$3.30 & 89.50$\pm$5.00 & 79.10$\pm$5.70 \\
\hline
Trans-SVNet \cite{gao2021trans} & \checkmark & 91.54$\pm$6.76 & \textbf{91.15$\pm$4.12} & 89.47$\pm$6.32 & 79.89$\pm$6.44 \\
\hline
SAHC \cite{ding2022exploring} & \checkmark & 91.80$\pm$8.10 & 90.30$\pm$6.40 & 90.00$\pm$6.40 & 81.20$\pm$5.50 \\
\hline
SKiT \cite{liu2023skit} & \checkmark & 93.40$\pm$5.20 & 90.90 & 91.80 & 82.60 \\
\hline
LoViT \cite{liu2025lovit} & \checkmark & 92.40$\pm$6.30 & 89.90$\pm$6.10 & 90.60$\pm$4.40 & 81.20$\pm$9.10 \\
\hline
\textbf{SPRMamba(Ours)} & \checkmark & \textbf{93.60$\pm$5.27} & 90.64$\pm$5.09 & \textbf{92.43$\pm$5.09} & \textbf{83.35$\pm$6.37} \\
\hline
TeCNO \cite{czempiel2020tecno} &  & 89.35$\pm$6.70 & 83.24$\pm$7.21 & 81.29$\pm$6.61 & 70.08$\pm$9.08 \\
\hline
Trans-SVNet \cite{gao2021trans} &  & 90.27$\pm$6.48 & 85.23$\pm$6.97 & 82.92$\pm$6.77 & 72.42$\pm$8.92 \\
\hline
SKiT \cite{liu2023skit} &  & 92.50$\pm$5.10 & 84.60 & 88.50 & 76.70 \\
\hline
LoViT \cite{liu2025lovit} &  & 91.50$\pm$6.10 & 83.10$\pm$9.30 & 86.50$\pm$5.50 & 74.20$\pm$11.30 \\
\hline
\textbf{SPRMamba(Ours)} &  & \textbf{93.12$\pm$4.58} & \textbf{89.26$\pm$6.69} & \textbf{90.12$\pm$5.61} & \textbf{81.43$\pm$6.90} \\
\bottomrule
\end{tabular}
}
\end{table}

Our comparison results are presented in Table \ref{tab2}. Our method outperforms the other methods in all evaluation metrics. Specifically, for average accuracy, SPRMamba attained an accuracy that exceeds the benchmark set by TeCNO by a margin of 3.77 pp. Moreover, our model demonstrates superior performance over LoViT, the leading model for phase recognition, by a difference of 1.62 pp in accuracy. It also exhibits more consistent performance, as evidenced by a reduced standard deviation in accuracy by roughly 2.12 pp in contrast to TeCNO. Additionally, for average precision, average recall, and average Jaccard, SPRMamba achieves the best performance with 89.26\%, 90.12\%, and 81.43\%, respectively. Beyond standard metrics, SPRMamba also proved to be more effective when evaluated against relaxed metrics.
\begin{figure*}[!t]
	\centerline{\includegraphics[width=\textwidth]{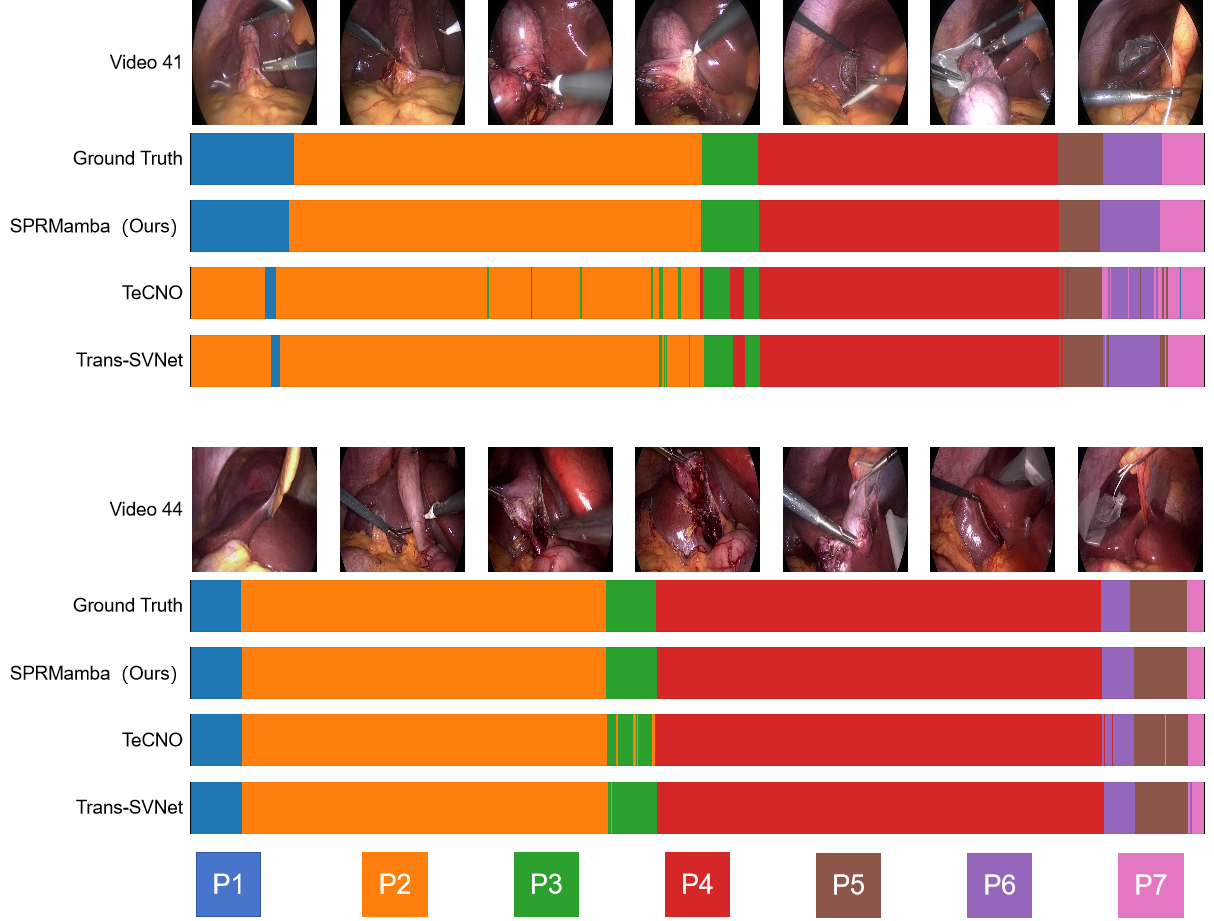}}
	\caption{Qualitative comparisons with some other methods on Cholec80 dataset. The following four rows show the ground-truth labels, the predictions by the proposed SPRMamba, the predictions by TeCNO, and the predictions by Trans-SVNet. P1 to P7 indicate the phase label.}
	\label{fig5}
\end{figure*}

Furthermore, to further demonstrate the effectiveness of our approach, Fig.\ref{fig5} presents a qualitative comparison between SPRMamba and two state-of-the-art methods (TeCNO \cite{czempiel2020tecno} and Trans-SVNet \cite{gao2021trans}) on representative samples from the Cholec80 test set. The results show that SPRMamba achieves more accurate phase recognition, particularly at transition boundaries where competing methods frequently misclassify frames, such as between phases P3/P4 and P5/P4. While TeCNO and Trans-SVNet exhibit noticeable instability with scattered misclassifications, our method maintains strong alignment with ground truth labels, reflecting its robustness in handling ambiguous frames and preserving temporal coherence. This improvement stems from SPRMamba's ability to effectively model spatial-temporal dependencies, reducing both boundary errors and over-segmentation artifacts observed in existing approaches. The visual analysis corroborates our quantitative findings, further validating the advantages of the proposed architecture.  
\subsection{Ablation Study}
We conducted a series of ablation experiments on the ESD385 dataset to validate the effectiveness of each component and parameter setting in our proposed method on the model.

\subsubsection{Scale Residual TranMamba} 
\begin{table}[ht]
\caption{Ablation study on the scale residual TranMamba. In the case of Transformer, we only use transformer branches instead of SRTM. In the case of Mamba, we only used the Mamba branch.}
\label{tab7}
\centering
\setlength{\tabcolsep}{3pt}
\resizebox{\textwidth}{!}{
\begin{tabular}{c|c|c|c|c}
\toprule
 & Accuracy(\%) & Precision(\%) & Recall(\%) & Jaccard(\%) \\
\hline
Transformer & 86.03$\pm$10.89 &	86.02$\pm$17.40 & 84.89$\pm$17.03 & 75.44$\pm$18.41 \\
\hline
Mamba & 86.91$\pm$12.06 & 86.59$\pm$17.24 &	86.05$\pm$16.74	& 77.19$\pm$18.42 \\
\hline
\textbf{SRTM(Ours)} & \textbf{87.70$\pm$11.03} & \textbf{86.84$\pm$16.07} & \textbf{86.89$\pm$15.64} & \textbf{77.74$\pm$17.60} \\
\bottomrule
\end{tabular}
}
\end{table}
Table \ref{tab7} shows the importance of the SRTM modules. Transformer and Mamba denote the use of transformer branches alone and Mamba branches alone, respectively. To keep the number of parameters constant, we still use \( {F} \in \mathbb{R}^{L \times 64} \) as input. SRTM improves the accuracy by 1.61\% and 0.73\% compared to Transformer and Mamba, respectively, demonstrating the crucial role of modeling short-term and long-term temporal context for ESD surgical phase recognition.

\subsubsection{Hierarchical Sampling Strategy} 
\begin{table}[ht]
\caption{Ablation study on the Hierarchical Sampling Strategy. In the case of STContext SRTM, we use two STContext SRTM blocks instead of a combination of STContext SRTM and LTContext SRTM. In the case of LTContext SRTM, we used two LTContext SRTM blocks.}
\label{tab6}
\centering
\setlength{\tabcolsep}{3pt}
\resizebox{\textwidth}{!}{
\begin{tabular}{c|c|c|c|c}
\toprule
 & Accuracy(\%) & Precision(\%) & Recall(\%) & Jaccard(\%) \\
\hline
baseline & 86.58$\pm$11.58 & 86.60$\pm$17.48 & 84.16$\pm$16.80 & 75.65$\pm$18.70 \\
\hline
STContext SRTM & 87.03$\pm$11.73 & 86.95$\pm$17.57 & 85.20$\pm$17.32	& 76.86$\pm$17.82 \\
\hline
LTContext SRTM & 87.38$\pm$11.15 & \textbf{87.27$\pm$17.20} & 85.42$\pm$16.43	& 77.07$\pm$17.89 \\
\hline
\textbf{SPRMamba(Ours)} & \textbf{87.70$\pm$11.03} & 86.84$\pm$16.07 & \textbf{86.89$\pm$15.64} & \textbf{77.74$\pm$17.60} \\
\bottomrule
\end{tabular}
}
\end{table}
To demonstrate the effectiveness of the Hierarchical Sampling Strategy, we designed four ablation experiments. We also configured the original SRTM framework matching the SPRMamba structure as a benchmark for comparison. As shown in Table \ref{tab6}, baseline achieves an average accuracy of 86.58\%, an average precision of 86.60\%, an average recall of 84.16\%, and an average Jaccard of 75.65\% for surgical phase recognition. The experimental results outperform the baseline modules regardless of whether the key modules are added individually or in combination with each other. First, we show the impact of combining STContext SRTM and LTContext SRTM in the LSTContext block shown in Fig \ref{framework}. We compare it with two variants where only STContext SRTM or LTContext SRTM is used. To keep the number of parameters constant, in these cases, we still use both SRTM blocks in LSTContext. The results show that combining the two SRTM blocks gives better results.

\subsubsection{Effect of Different Number Of Layers}
\begin{table}[ht]
\caption{Ablation study on different number of layers.}
\label{layers}
\centering
\setlength{\tabcolsep}{3pt}
\resizebox{\textwidth}{!}{
\begin{tabular}{c|c|c|c|c}
\toprule
Layers & Accuracy(\%) & Precision(\%) & Recall(\%) & Jaccard(\%) \\
\hline
8 & 86.91$\pm$11.17 & 86.88$\pm$17.37 & 84.87$\pm$17.28 & 76.31$\pm$18.16 \\
\hline
9 & 86.94$\pm$10.78 & 87.37$\pm$17.82 & 84.96$\pm$16.93 & 76.76$\pm$17.77 \\
\hline
\textbf{10} & \textbf{87.70$\pm$11.03} & 86.84$\pm$16.07 & \textbf{86.89$\pm$15.64} & \textbf{77.74$\pm$17.60} \\
\hline
11 & 87.50$\pm$10.75 & \textbf{87.49$\pm$17.04} & 85.68$\pm$17.37 & 77.41$\pm$18.40 \\
\hline
12 & 87.64$\pm$9.83 & 87.17$\pm$17.49 & 85.49$\pm$16.01 & 77.11$\pm$17.77\\
\bottomrule
\end{tabular}
}
\end{table}
The ablation study investigating the impact of varying the number of layers in the LSTContext blocks is presented in Table \ref{layers}. The results demonstrate that the model's performance is generally robust to changes in layer depth, with all configurations achieving competitive metrics. Notably, the 10-layer configuration achieved the highest accuracy, recall, and Jaccard, while the 11-layer variant yielded the best precision. The marginal differences between the 10-, 11-, and 12-layer models suggest that the architecture saturates in performance beyond 10 layers, with diminishing returns for additional depth. Based on these findings, we selected N=10 for the ESD385 datasets, respectively, balancing computational efficiency and task-specific requirements. 
\subsubsection{Effect of Different Number Of Stages}
\begin{table}[ht]
\caption{Ablation study on different number of stages.}
\label{stages}
\centering
\setlength{\tabcolsep}{3pt}
\resizebox{\textwidth}{!}{
\begin{tabular}{c|c|c|c|c}
\toprule
Stages & Accuracy(\%) & Precision(\%) & Recall(\%) & Jaccard(\%) \\
\hline
1 & 87.16$\pm$11.07 & 86.86$\pm$17.51 & 84.85$\pm$18.18 & 76.36$\pm$18.79 \\
\hline
2 & 86.68$\pm$11.22 & 86.77$\pm$17.45 & 85.16$\pm$17.12 & 76.43$\pm$18.38 \\
\hline
3 & 87.31$\pm$10.40 & 86.85$\pm$18.17 & 85.01$\pm$17.03 & 76.51$\pm$18.18\\
\hline
\textbf{4} & \textbf{87.70$\pm$11.03} & 86.84$\pm$16.07 & \textbf{86.89$\pm$15.64} & \textbf{77.74$\pm$17.60} \\
\hline
5 & 87.14$\pm$10.26 & \textbf{87.18$\pm$16.57} & 85.66$\pm$16.01 & 77.20$\pm$16.64 \\
\bottomrule
\end{tabular}
}
\end{table}
As shown in Fig.\ref{framework}, we use four stages of LSTContext blocks, each of them with 10 layers. To assess the influence of stage depth, we conducted an ablation study, with the results summarized in Table \ref{stages}. The analysis reveals that employing multiple stages effectively mitigates over-segmentation errors while substantially enhancing both the Accuracy and Jaccard compared to using only single-stage. Performance metrics consistently improve as the number of stages increases up to four, achieving optimal accuracy, recall, and Jaccard. However, expanding to five stages leads to a decline in performance, indicating the onset of overfitting. These findings validate the selection of four stages as the optimal balance between model complexity and generalization capability.
\subsubsection{Effect of Different Value of W and G}
\begin{table}[ht]
\caption{Ablation study on different parameters W and G.}
\label{tab5}
\centering
\setlength{\tabcolsep}{3pt}
\resizebox{\textwidth}{!}{
\begin{tabular}{c|c|c|c|c|c}
\toprule
W & G & Accuracy(\%) & Precision(\%) & Recall(\%) & Jaccard(\%) \\
\hline
8 & 64 & 87.53$\pm$11.10 & \textbf{87.73$\pm$16.56} & 85.39$\pm$16.21 & 77.41$\pm$17.16 \\
\hline
16 & 64 & 87.37$\pm$9.63 & 86.56$\pm$16.23 & 86.23$\pm$16.30 & 77.35$\pm$17.17 \\
\hline
32 & 64 & 86.91$\pm$11.52 & 87.62$\pm$17.82 & 85.10$\pm$15.87 & 76.80$\pm$18.23 \\
\hline
\textbf{64} & \textbf{64} & \textbf{87.64$\pm$9.83} & 86.72$\pm$16.54 & \textbf{86.76$\pm$15.66} & \textbf{77.51$\pm$17.83} \\
\hline
64 & 32 & 87.33$\pm$10.41 & 86.50$\pm$16.74 & 85.62$\pm$16.32 & 76.85$\pm$17.65 \\
\hline
64 & 16 & 87.08$\pm$10.98 & 87.09$\pm$17.19 & 85.15$\pm$16.42 & 76.77$\pm$17.91\\
\hline
64 & 8 & 87.11$\pm$10.76 & 86.35$\pm$18.76 & 85.88$\pm$17.40 & 76.60$\pm$18.76\\
\bottomrule
\end{tabular}
}
\end{table}
To further explore the effect of the Hierarchical Sampling Strategy with different W and G on the model performance, we conducted the experiments shown in Table \ref{tab5}. The parameter W controls the size of the local window and the parameter G controls the range of the long-term temporal context modeling. The experimental results show that the performance of the models (Accuracy and Jaccard) increases as G increases until the optimal performance is reached for G=64. The performance fluctuates as the size of the local window W becomes smaller, but W=64 works best.

\subsubsection{Effect of Using Convolutions}
\begin{table}[ht]

\caption{Ablation study on using convolutions.}
\label{tab4}
\centering
\setlength{\tabcolsep}{3pt}
\resizebox{\textwidth}{!}{
\begin{tabular}{c|c|c|c|c}
\toprule
 & Accuracy(\%) & Precision(\%) & Recall(\%) & Jaccard(\%) \\
\hline
Conv & 85.97$\pm$11.48 & 85.75$\pm$17.70 & 83.96$\pm$18.06 & 74.74$\pm$18.92 \\
\hline
w/o Conv & 86.60$\pm$10.95 & 85.71$\pm$17.94 & 84.99$\pm$17.56 & 75.46$\pm$18.87 \\
\hline
\textbf{Dilation} & \textbf{87.64$\pm$9.83} & \textbf{86.72$\pm$16.54} & \textbf{86.76$\pm$15.66} & \textbf{77.51$\pm$17.83} \\
\bottomrule
\end{tabular}
}
\end{table}
To explore the impact of dilated convolution in LSTContext blocks on model performance. We compared 1D dilated convolution with 1D convolution using the same size but without dilation factor and without 1D convolution. The results show that dilated convolution has a significant impact on performance.

\section{Disscussion}
The accurate recognition of surgical phases in ESD within CAS systems is crucial for enhancing surgical efficiency, minimizing patient complications, and providing valuable training material for novice endoscopists. However, considering the duration and complexity of the ESD process, the challenge of handling long sequences of video frames and effectively modeling temporal context with limited computing resources remains significant. 

To address these challenges, we propose SPRMamba, a Mamba-based framework for online ESD surgical phase recognition. Our approach leverages the Scale Residual TranMamba module to effectively model both long-term and short-term temporal contexts, offering superior temporal modeling capabilities compared to traditional methods. Additionally, we introduce a novel Hierarchical Sampling Strategy to mitigate the computational resource constraints, making the framework feasible for real-time surgical phase recognition. Both quantitative and qualitative results form our analytical anlation experiments conducted on the ESD385 dataset demonstrate: 
\begin{figure*}[!t]
	\centerline{\includegraphics[width=\textwidth]{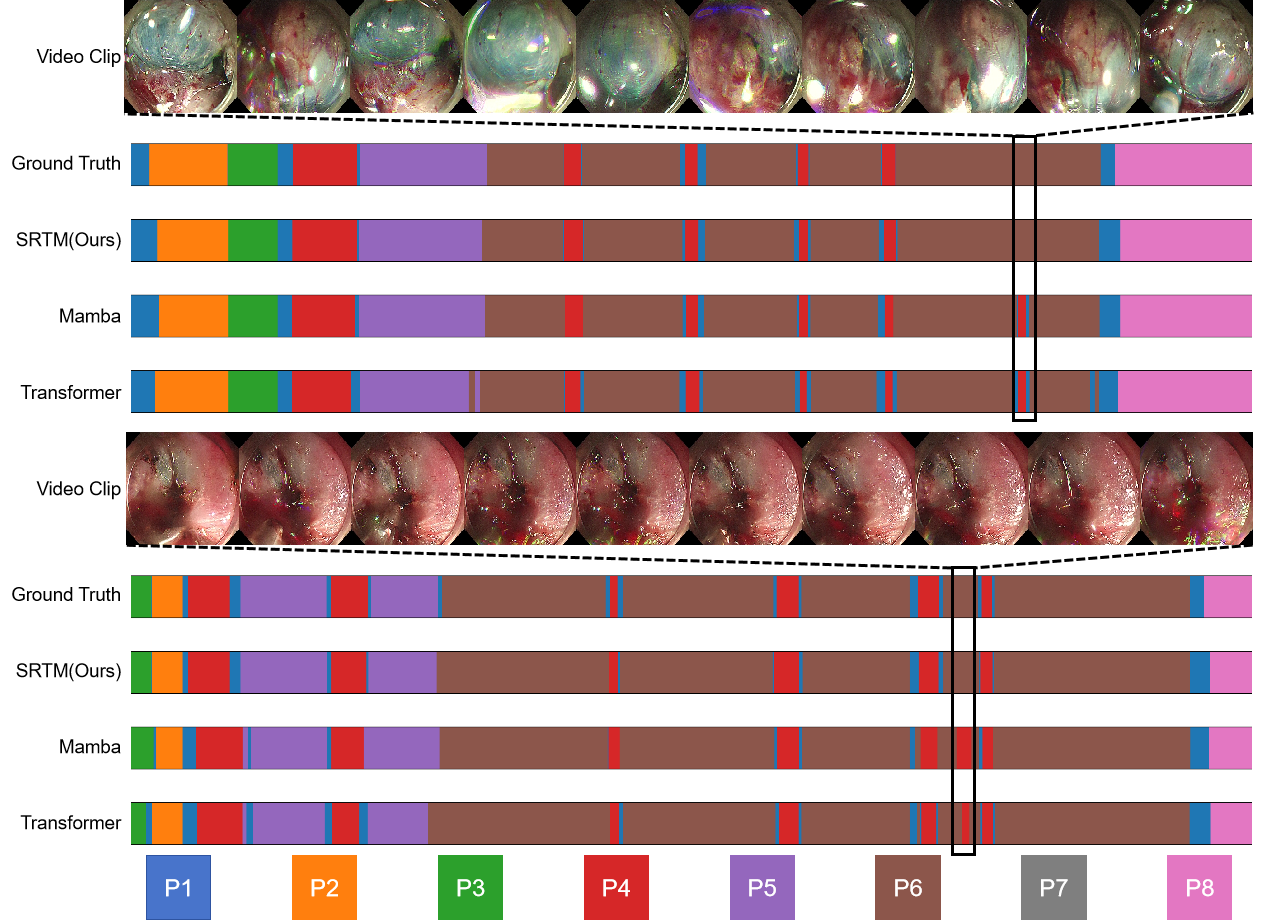}}
	\caption{Qualitative comparisons with SRTM, Mamba, and Transformer on the ESD385 dataset. The first row displays video clip corresponding to the black box containing segments, indicating examples of errors in SOTA methods other than ours. The following four rows show the ground truth, the predictions by the proposed SPRMamba, the predictions by using Mamba only, and the predictions by using the Transformer only. P1 to P8 indicate the phase label.}
	\label{ablation}
\end{figure*}

\textbf{(1) Long-Term Temporal Modeling with Linear Complexity:} By integrating Mamba’s state-space architecture, SPRMamba achieves superior long-range dependency modeling (capturing phases up to 20+ minutes) at 27\% lower FLOPs than Transformer-based methods (SAHC). This linear complexity (\(O(L)\) vs. \(O(L^2)\)) is critical for real-time ESD analysis, where traditional approaches like MS-TCN \cite{furube2024automated} and Trans-SVNet \cite{gao2021trans} fail to balance accuracy and speed (Table~\ref{tab1}).

\textbf{(2) Fine-Grained Phase Transition Resolution via SRTM:} The Scale Residual TranMamba (SRTM) module uniquely addresses ESD's low inter-phase variance (e.g., Incision and ESD phases) by combining Mamba’s global context with Transformer-like local attention. This hybrid design reduces misclassifications in short-duration phases (e.g., Preparation) compared to pure Mamba architectures, as validated in Fig.~\ref{ablation}.

\textbf{(3) Hierarchical Sampling Strategy for Clinical Feasibility:} The proposed Hierarchical Sampling Strategy outperforms direct long sequence processing and mitigates the secondary complexity typically associated with Transformer models, enhancing the efficiency of temporal context modeling and improving overall recognition performance. Crucially, it supports adaptive temporal modeling lengths, accommodating variability in ESD workflows across surgeons.  

Finally, compared with the SOTA method, our proposed method not only achieves better results in ESD surgical phase recognition, but also demonstrates robustness for other surgical tasks. Specifically, our method achieves an average accuracy of 87.64\%, an average precision of 86.72\%, an average recall of 86.76\%, and an average Jaccard of 77.51\% on the ESD dataset, outperforming the next best method by significant margins. Additionally, validation on the cholecystectomy Cholec80 dataset further confirms our method's robustness, achieving superior results in most metrics compared to the SOTA methods \cite{twinanda2016endonet, jin2020multi, tao2023last, jin2017sv, czempiel2020tecno, jin2021temporal, gao2021trans, ding2022exploring}, except for precision. The qualitative results, as illustrated in Fig.\ref{fig4} and Fig.\ref{fig5}, further substantiate the effectiveness and reliability of our approach.

Although our method shows high surgical phase recognition performance in ESD and cholecystectomy videos, it has some limitations. First, this study primarily focuses on ESD procedures, and while we validated the robustness of our method on the Cholec80 dataset, further experiments are needed to assess its generalizability across a broader range of surgical tasks. In the future, we plan to extend our method to other surgical video analysis tasks, potentially incorporating other innovations such as uncertainty analysis to enhance the robustness and accuracy of SPRMamba across different surgical settings. Second, the dataset used in this study was limited in size and diversity, which may affect the model's performance in more diverse clinical settings. Future work will focus on expanding the dataset to include multicenter data and a wider variety of surgical procedures.

\section{Conclusion}
\label{sec:conclusion}
In this paper, we address the critical need for accurate surgical phase recognition in Endoscopic Submucosal Dissection, a key procedure for treating early-stage gastrointestinal cancers. Achieving precise real-time phase recognition in ESD is essential for improving surgical outcomes and efficiency. However, traditional SPR methods face significant challenges, particularly in effectively capturing the temporal context over extended surgery durations and managing the high computational demands of video analysis required for real-time applications. To overcome these challenges, we propose SPRMamba, a novel framework designed to enhance the accuracy and efficiency of ESD surgical phase recognition. Specifically, our method proposes the Scale Residual TranMamba module, which combines the ability of Mamba to extract the long-term temporal context with the ability of the Transformer to extract the short-term temporal context and excels in capturing complex temporal relationships in surgical videos. In addition, considering the real-time requirements of surgical phase recognition, a temporal sampling strategy is designed to optimize computational resources by efficiently modeling both short-term and long-term temporal contexts. Extensive experiments demonstrate that SPRMamba not only outperforms current state-of-the-art methods in accuracy and robustness but also significantly reduces the computational burden. In the future, our approach has the potential to advance the field of surgical video analysis, offering a valuable tool for both clinical practice and surgical education.

\section*{Acknowledgment}
This work was supported by the Joint Laboratory of Intelligent Digestive Endoscopy between Shanghai Jiaotong University and Shandong Weigao Hongrui Medical Technology Co., Ltd. and the National Science Foundation of China under Grant 62103263.

\bibliographystyle{elsarticle-num} 
\bibliography{reference}
\end{document}